\documentclass[lettersize,journal]{IEEEtran}
\usepackage{amsmath,amsfonts}
\usepackage{algorithmic}
\usepackage{algorithm}
\usepackage{array}
\usepackage[caption=false,font=normalsize,labelfont=sf,textfont=sf]{subfig}
\usepackage{textcomp}
\usepackage{stfloats}
\usepackage{url}
\usepackage{verbatim}
\usepackage{graphicx}
\usepackage{cite}
\usepackage{xcolor}
\usepackage{makecell}
\usepackage{orcidlink}
\usepackage{glossaries}
\usepackage{multirow}
\usepackage{booktabs}
\usepackage[table,xcdraw]{xcolor}
\usepackage{siunitx}
\usepackage{graphicx}
\usepackage{float}

\newcommand{\teleopcolorcell}[3]{%
  \cellcolor[HTML]{#1}\ensuremath{#2 \pm #3}%
}

\newcommand{\R}[2]{{}^{\mathrm{#1}}\mathbf{R}_{\mathrm{#2}}}

\newacronym{fbg}{FBG}{Fiber Bragg Grating}
\newacronym{dvrk}{dVRK}{da Vinci Research Kit}
\newacronym{cwru}{CWRU}{Case Western Reserve University}
\newacronym{mtm}{MTM}{Master Tool Manipulator}
\newacronym{psm}{PSM}{Patient Side Manipulator}
\newacronym{jnd}{JND}{Just Noticeable Difference}
\newacronym{ramis}{RAMIS}{Robot-Assisted Minimally Invasive Surgery}
\newacronym{dof}{DoF}{degrees of freedom}
\newacronym{nrmse}{NRMSE}{Normalized Root Mean Squared Error}
\newacronym{rmse}{RMSE}{Root Mean Squared Error}
\newacronym{lstm}{LSTM}{Long Short-Term Memory}
\newacronym{fcn}{FCN}{Fully Connected Neural}

\begin{document}

\title{Shaft-integrated Force Sensing with Transformer-based Dynamics Compensation for Telesurgery}

\author{
Shuyuan Yang$^{1\dagger}$\,\orcidlink{0009-0001-1472-6529},~\IEEEmembership{Student Member,~IEEE},
Grant Boone$^{2\dagger}$\,\orcidlink{0009-0009-9712-0247},
Timo Markert$^{3,4}$\,\orcidlink{0000-0001-9121-671X},
Sebastian Matich$^3$, \\
Andreas Theissler$^5$\,\orcidlink{0000-0003-0746-0424}, 
Martin Atzmueller$^{4,6}$\,\orcidlink{0000-0002-2480-6901},
and Zonghe Chua$^1$\,\orcidlink{0000-0002-6101-5412},~\IEEEmembership{Member,~IEEE}

\thanks{Manuscript received January 22, 2026; revised April 20, 2026; accepted May 22, 2026. \textit{(Corresponding author: Zonghe Chua)}}
\thanks{$^1$S.Yang, and Z.Chua are with the Department of Electrical, Computer, and Systems Engineering, Case Western Reserve University, Cleveland, OH 44016, USA. \texttt{\{sxy841,zxc703\}@case.edu}}
\thanks{$^\dagger$These authors contributed equally to this work.}
\thanks{$^2$G.Boone was with the Department of Mechanical and Aerospace Engineering, Case Western Reserve University, Cleveland, OH 44106, USA. \texttt{gjb74@case.edu}}
\thanks{$^3$T. Markert, and S. Matich are with Resense GmbH, Germany.}
\thanks{$^4$T. Markert and M. Atzmueller are with the Semantic Information Systems Group, Osnabrück University, Osnabrück, Germany.}%
\thanks{$^5$A. Theissler is with Justus Liebig University, Giessen, Germany.}%
\thanks{$^6$M. Atzmueller is also with the German Research Center for Artificial Intelligence (DFKI), Osnabrück, Germany.}%
}

\markboth{IEEE Transactions on Medical Robotics and Bionics. Preprint Version. }{Yang \MakeLowercase{\textit{et al.}}: Shaft-integrated Force Sensing with Transformer-based Dynamics Compensation for Telesurgery}


\maketitle

\begin{abstract}
\gls{ramis} enhances surgeon dexterity, with newer platforms leveraging haptic feedback to further improve performance. Such force information has broader potential to inform performance assessment, tactile localization, and surgical autonomy. This motivates the need for accessible approaches to integrating force sensing into \gls{ramis} tools.
This work presents a method for integrating a six-axis commercial force sensor into the distal end of a standard cable-driven surgical instrument, enabling end-effector force measurement while preserving the original mechanical functionality of the device. The proposed design emphasizes reproducibility and accessibility for research applications, requiring no specialized manufacturing tools. A transformer neural network integrates force sensor measurements with robot state information to aid estimation of applied forces at the end-effector, compensating for internal cable forces arising from actuation. 
Our proposed approach achieved normalized errors below 6\%, and generalized to unseen conditions better than purely proximal data-driven sensing approaches. High internal cable forces caused sensor saturation and reduced axial force observability, which can degrade performance along the tool's major axis and under higher load conditions.
Given current levels of performance, the balance of system integrability and performance enables applications and research into timely topics of haptic feedback, skill assessment, and force-informed autonomy in \gls{ramis}. Videos and code are available at \url{https://enhanced-telerobotics.github.io/shaft_force_sensing}.
\end{abstract}


\section{Introduction}
\IEEEPARstart{R}{AMIS}~\cite{what_is_ramis} is a surgical approach that combines robotic systems with minimally invasive techniques, allowing surgeons to perform procedures through small incisions, using teleoperation to control dexterous instruments. 
During teleoperation, the surgeon interacts with the environment indirectly through the robotic system, without direct physical contact. Therefore by default, haptic perception is not directly provided. Instead, it must be enabled through force sensing and feedback, or be inferred indirectly through visual observation.

\subsection{Benefits of Haptic Information in Telesurgery}
Haptic feedback in robotic surgical instruments has been shown to provide several important benefits during procedures. 

First, haptic feedback reduces both the peak and average forces applied by surgeons to tissue, which has the potential to lower the risk of unintentional injury~\cite{benefits_of_robot_surgury,benefits_of_robot_surgury_2}. In addition, access to force feedback can improve task efficiency, with reductions in task completion time and increases in procedural success rate~\cite{benefits_of_robot_surgury}. 

\begin{figure}[t]
    \centering
    \includegraphics[width=1\linewidth]{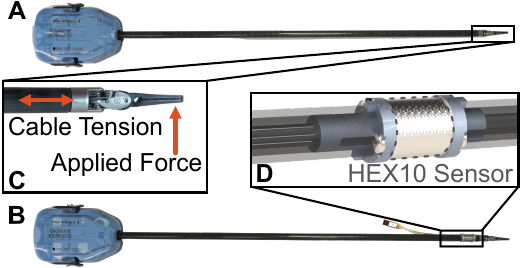}
    \caption{(A) EndoWrist Cadiere Forceps tool with (B) shaft-integrated sensor assembly. (C) Side loading and direction of induced cable tension force in the axial direction. (D) CAD model of HEX10 sensor interfaced with EndoWrist axial and shaft control rods passing through center bore.}
    \label{fig:tool_sensor}
\end{figure}


%
Second, force information plays a key role in the development of future advancements in \gls{ramis}. This includes autonomous or semi-autonomous processes such as tissue palpation for property estimation~\cite{haptic_palpation2}, enabling fast and accurate segmentation of underlying tumors from other tissue~\cite{haptic_palpation2}. More recently, tool tip force sensing has been shown to improve performance in learning-based dexterous manipulation, by both directly informing the autonomous policy~\cite{hou2025adaptive,haptic_palpation, lee2020making}, and also expert training demonstrations~\cite{cuan_leveraging_2024}. In particular, Abdelaal et al. demonstrated that a force-aware autonomous policy was on average, less forceful, and three times more successful than a force-agnostic variant in basic telesurgical retraction~\cite{haptic_palpation}. 

\subsection{Low Availability of Integrated Force Sensing}
Manufacturing and integrating miniature multi-axis force sensors into the cable-driven \gls{ramis} tools is complex. Additionally, the high gear ratios and non-linear dynamics make non-collocated sensing of end-effector forces difficult. Consequently, these factors hinder robust, generalizable solutions. Except for the state-of-the-art da Vinci 5, other commercial \gls{ramis} systems lack force-sensing features. Research platforms such as the da Vinci Research Kit also do not have native force-sensing capability. Therefore, researchers developing novel haptic feedback approaches, or collecting data to train and research autonomous surgical manipulation policies, must develop their own force-sensing approaches. Alternatively, they resort to embedding the sensor in the manipulation environment, limiting their ability to extend and evaluate their approaches to bimanual manipulations~\cite{yang2024vision,haptic_palpation}. Open-source approaches also remain limited, and require fabricating custom printed circuit boards, and careful calibration~\cite{chua_3dof_intro_motivation}. 

\subsection{Contributions}
In this work, we adopt an integrated hardware and software approach that standardizes and simplifies force sensing in \gls{ramis} tools, thereby improving accessibility to force sensing as a research tool, with a forward view towards informing design in commercial settings. Specifically, we present:

\begin{itemize}
    \item An accessible and replicable method for integrating a tubular six-axis force/torque sensor into the \gls{ramis} tool shaft for compatibility with existing research platforms;
    \item a transformer time-series regression model for force-informed dynamics compensation;
    \item a calibration protocol and dataset for fitting the dynamics compensation model;
    \item experimental validation of performance sensitivity to model inputs, overload characteristics;
    \item and a transfer learning protocol for generalization to new robot configurations.
\end{itemize}
\vspace{1.25em}

\section{Background}
\subsection{Challenges of Shaft-mounted Distal Force Sensing}\label{sec:technical_challenges}

For \gls{ramis} tools such as the da Vinci EndoWrists in Fig.\,\ref{fig:tool_sensor}A, the wrist pitch and both jaw joints are actuated via control pulleys in the base. The pretension of the control cables maintain the structural integrity of the assembly, holding the shaft, wrist, and base under sustained axial tension. This design thus has highly non-linear dynamics arising from internal friction forces, cable backlash, and hysteresis. Although these forces can be partially compensated using dynamic modeling, they are highly dependent on their spatiotemporal initial configurations~\cite{okamura_haptics_sensor_placement}. Rotation about the axial direction, i.e. roll, further introduces non-linearities as the control cables and rods are twisted around each other, which can further increase friction and create internal compression forces.

Additionally, the length of the jaw component of certain end-effectors like the Cadiere Forceps in Fig.\,\ref{fig:tool_sensor}A-B is significantly longer than the radius of the pulleys at the wrist, as illustrated in Fig.\,\ref{fig:tool_sensor}C. Because of this, forces applied to the tip of the end-effector can create large moments about the wrist pulleys, which must be resisted by the control pulleys using high cable tensions, resulting in high internal axial forces inside the shaft~\cite{blumenkranzForceTorqueSensing2015}. These forces are significantly larger than the interaction forces, thus limiting the practical range of the sensors, and reducing the signal-to-noise ratio.

\subsection{Hardware-based Force Sensing}\label{sec:bg_hardware}

A variety of hardware approaches utilizing different sensing modalities and sensing locations have been developed for telesurgical end-effector force sensing. Modalities include strain gauges~\cite{proximal_shaft_force_sense,strain_gauge_3dof}, capacitance~\cite{kim_sensorized_2018}, and light intensity~\cite{hosseinabadi2020ultralow}, or frequency modulation~\cite{x-perce_fbg}. Strain gauge, capacitance, and piezoelectric approaches are cost-efficient, but are susceptible to disturbances from electrocautery, temperature variations, and can be challenging to sterilize~\cite{background_force_sense_methods,background_force_sense_methods_2}. Optical approaches, particularly fiber-Bragg gratings can mitigate these issues, but require more costly instrumentation~\cite{background_force_sense_methods,background_force_sense_methods_2}. Integration locations include the instrument base~\cite{base_force_sense}, as well as the proximal~\cite{proximal_shaft_force_sense}, and distal~\cite{distal_shaft_force_sense} ends of the instrument shaft. Compared to shaft integrations, instrument base integrations are easier to sterilize, but are subject to external disturbances and noise from aforementioned tool transmission dynamics, and surgical port forces.  

\subsection{Software-based Force Estimation}\label{sec:bg_hardware}

Model-based and data-driven force estimation has been applied to derive end-effector interaction forces \cite{background_force_sense_methods_2}, to reduce sensor integration complexity. In model-based approaches, encoder-based joint positions and current-based torque measurements are combined to estimate end-effector forces using dynamics modeling~\cite{wang_convex_2019, fontanelli_modelling_2017}, or disturbance observers~\cite{yilmaz2023sensorless}. However, analytical modeling of the cable-driven surgical instruments is challenging due to the many sources of non-linearities, and especially with the coupled \gls{dof} in the wristed instruments~\cite{background_force_sense_methods_2}. Data-driven, machine learning approaches are increasingly used to approximate the dynamics~\cite{yilmaz_neural_2020,yilmaz2023sensorless,nowakowski2024learning}, with many adopting time-series methods to handle hysteresis and viscous phenomena~\cite{yilmaz2023sensorless,nowakowski2024learning}. Besides these intrinsic estimation approaches, extrinsic, vision-based techniques can also be employed. Here, external end-effector forces are derived using visual feedback of the tissue deformation through cameras~\cite{marban_recurrent_2019,chua2021toward,lee2023learning,yang2024vision}. Vision-based techniques have two key limitations. First, they can be sensitive to changes in the camera's view, such as instrument obstructions and changes in lighting. Secondly, in vision-based approaches, the estimation update rate cannot exceed the camera frame rate, which is often limited to 30 Hz. 

In more recent work, an integrated hardware-software approach has been applied to cancel out internal disturbances arising from the dynamics of a cable-actuated instrument. This integrates two sensors, one at the distal end of the shaft, and one at the proximal end, and uses a neural network to fuse the differential information and estimate the end-effector force~\cite{resense_differential_force_compensation}. 

Overall, existing research still leaves several gaps:
\begin{itemize}
    \item Many approaches are difficult to replicate as they require fabricating custom electromechanical transducers;
    \item external hardware modifications alter the design of the original \gls{ramis} system, particularly tool tip geometry;
    \item software solutions rely solely on complex modeling while still offering limited accuracy gains.
\end{itemize}

\section{Design Objectives}
\label{sec:design_obj}
We address the above limitations by targeting the following design objectives:

\begin{figure*}[t]
    \begin{minipage}{5.5in}
    \centering
    \includegraphics[width=1\linewidth]{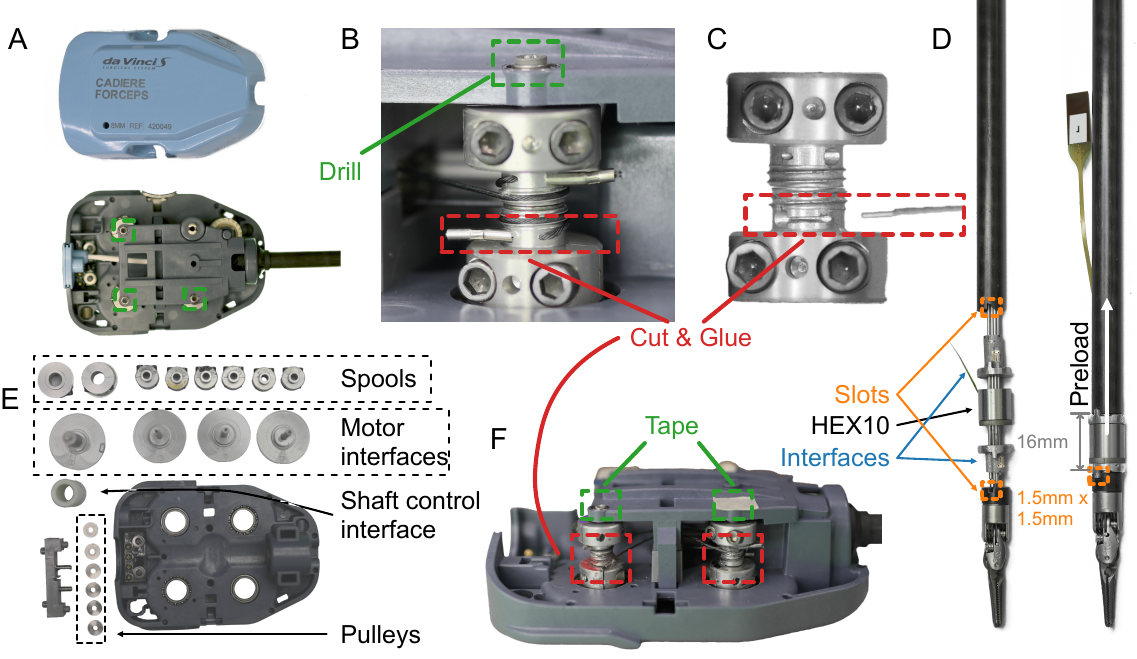}
    \end{minipage}
    \begin{minipage}{1.4in}
    \vspace{5em}
    \caption{
    EndoWrist modification process. (A) Non-destructive removal of the case and upper bearing assembly. (B) Drilling the motor interface to remove the bearing and spools. (C) Cutting a slot in the spool to free the control cables. (E) Removing all spools and cables to free the shaft. (D) Removing and shortening the shaft to accommodate the HEX10 and its interfaces; rewiring the control cables through the shaft and reattaching them to the spools. (F) Securing and retensioning the assembly cables to the spools with cyanoacrylate and tape bearings at the motor interface to prevent loosening. 
    }
    \label{fig:modification_summary}
    \end{minipage}
\end{figure*}

\textit{Accessible Modification Process:} A key objective of this work was to develop an accessible and reproducible method for integrating a sensor into a standard cable-driven \gls{ramis} instrument. We operationalize accessibility in terms of costs, time, and infrastructure using the following criteria: (a) integration cost below \$100 USD, excluding baseline sensor cost; (b) sub-day active manufacturing time, i.e., $<$ 8 hours; and (c) requiring only manufacturing techniques available to a typical university lab (i.e., 3D printing and CNC machining). We rely on using software-based data-driven approaches to compensate for any noise and uncertainties resulting from this approach. This design feature is particularly valuable for researchers looking to integrate force sensing into their \gls{ramis} systems (e.g. \acrlong{dvrk}~\cite{dvrk_main}, and the Raven~\cite{raven}), to develop new force-informed autonomous policies.  

\emph{Preservation of Function and Structural Integrity:} The modification process does not compromise the original mechanical performance of the \gls{ramis} instrument. After sensor integration, the device retains its full range of motion, dexterity, and original control method while maintaining structural stability under typical loads. Ensuring functional equivalence to the unmodified device as well as mechanical robustness is critical to validating the instrument as a reliable research platform.

\emph{Accurate Force Sensing:} Accuracy metrics for force sensing are informed by the types of targeted application. Firstly, given our focus on haptic feedback, we target the average Weber Fraction for kinesthetic forces for the human hand, which is 12.5\%~\cite{just_noticable_difference}. This can be mapped to targeting an equivalent \acrlong{nrmse}~\cite{cannula_force_sense}. Second, even if absolute accuracy is not achieved, the ability to track force variations would enable measurements to inform human or autonomous decisions as demonstrated by the utility of image-based tactile feedback systems for autonomy~\cite{ablett2024multimodal}, and haptic feedback via sensory substitution~\cite{quek_sensory_2015,vuong2025effects}. Because of this, we also calculate the $R^2$ correlation of the predicted force to the ground truth data for each axis, targeting a correlation above 90\%.

\section{Hardware Design}

In our design, we integrated a HEX10 6-axis force sensor (Resense GmbH, Germany) into the shaft of an off-the-shelf Cadiere Forceps EndoWrist tool (Intuitive Surgical Inc., Sunnyvale, CA). The HEX10 was chosen because of its tubular profile, with the outer diameter only 1.6 mm wider than an EndoWrist shaft. Crucially, it has an inner diameter that is large enough to allow the end-effector control rods to pass through and move. Additionally, the HEX10 sensor has been applied to prior research into force sensing for the application of haptics to surgical tasks~\cite{resense_palpation_2022}.  

\subsection{Shaft Interface}

In order to integrate the HEX10 sensor into the shaft of the EndoWrist, we designed several components to interface between the shaft and the sensor. The interface components needed to allow the continued function of the control cables, accommodate the M1.4 screws that are compatible with the HEX10's interface plates, resist applied torques, and ensure the structural rigidity of the completed assembly. The design of the interfacing components is presented in Fig.\,\ref{fig:tool_sensor}C. They consist of two flanged cylindrical components that would screw into either side of the sensor. The interface components slot into the shaft on either side of the EndoWrist with a depth of 10\,mm. Tabs on the flange of the interface were designed to mate with slots cut manually into the shaft of the EndoWrist, as shown in Fig.\,\ref{fig:tool_sensor}B. This feature was included to resist applied torques, and prevent rotations about the long axis of the tool. The interface components were manufactured out of aluminum using commercial metal sintering services from Craftcloud. We note that a PETG version can also be 3D-printed using a desktop fused-deposition modeling printer, at the expense of a lower usable force range.

\subsection{Shaft Integration}

Modification of a standard EndoWrist tool began with disassembly for shaft-level sensor integration. EndoWrist instruments are typically held together by actuation cable-and-rod sub-assemblies that drive the distal end-effector joints through pulleys in the proximal housing (Fig.\,\ref{fig:modification_summary}A). These transmissions provide both actuation and the pretension that maintains structural integrity. Following Fig.\,\ref{fig:modification_summary}B-C, slots were cut into the spools to release each cable without permanently severing it. After releasing the cables from their spools, the shaft was removed and shortened to create space for the sensor (Fig.\,\ref{fig:modification_summary}D). The sensor-integrated shaft was then reassembled with the proximal housing, and each cable was routed back through its spool slot and rewound. The spool slots were sealed with cyanoacrylate adhesive (Fig.\,\ref{fig:modification_summary}F). Once reassembled, the cables were retensioned to minimize backlash while using the HEX10 force reading to monitor preload. The target preload was defined as  $\sim\SI{8}{\newton}$ for the axial force. This balanced preserving the sensor's available range and the instrument's structural and functional properties. A video summarizing the process is provided on the project website.

\section{Data-driven Force Estimation Approach}

To address the challenge of decoupling the non-linear internal forces from the end-effector interaction forces in the force sensor measurements, we implemented a data-driven spatiotemporal modeling approach. The end-effector interaction forces are estimated using both the six-axis measurements from the HEX10, and the joint state information from the \gls{dvrk}. The torque estimates from the joint state provide information about the internal cable forces, while the joint positions, allow the model to condition estimates on the joint configuration. Prior work suggests that such models would be effective for force estimation in cable-driven systems like the EndoWrist~\cite{data_driven_modeling_endowrist}.

\subsection{Model Architecture}

\begin{figure}[t]
    \centering
    \includegraphics[width=0.8\linewidth]{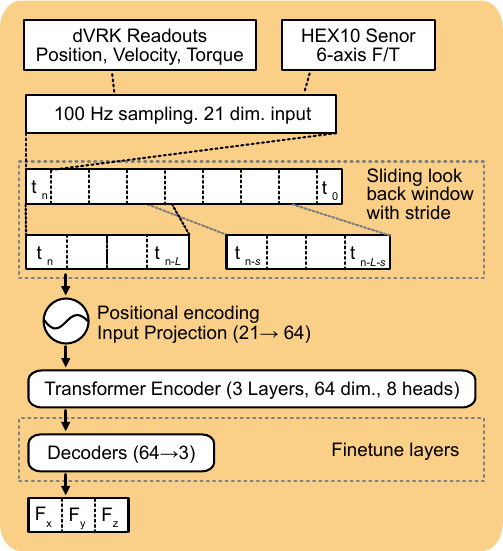}
    \caption{Transformer model with a temporal look-back window and striding on the input.}
    \label{fig:model_arch}
\end{figure}

To enable modeling of spatiotemporal dependencies, we used a transformer neural network to perform time-series regression over a sequence of input features. The transformer was implemented as 3 encoder layers with 8 attention heads and a model dimension of 64. A two-layer \acrlong{fcn} network were used to decoding a final three-dimensional force estimate. An input sequence consisted of $L$ consecutive previous samples. Consecutive sequences were sampled with a temporal separation of $s$. This temporal processing of the inputs is illustrated graphically in Fig.\,\ref{fig:model_arch}. The model's output was a 3-dimensional vector of the external Cartesian forces at the tool tip, specified in the HEX10's reference frame. 

\begin{table}[t]
\centering
\caption{Model inputs and outputs by source.}
\label{tbl:model_io_sources}
\setlength{\tabcolsep}{10pt}
\renewcommand{\arraystretch}{1.2}
\begin{tabular}{ccc}
\toprule
\textbf{Source} & \textbf{Axes} & \textbf{Signals (Dim.)} \\
\midrule
\multicolumn{3}{c}{\textbf{Input Data}} \\
\midrule
\multirow{7}{*}{dVRK PSM} 
 & Jaw         & $q_{7}$, $\dot{q}_7$, $\tau_7$ (3) \\
 & Roll        & $q_{4}$, $\dot{q}_4$, $\tau_4$ (3) \\
 & Wrist Pitch & $q_6$, $\dot{q}_6$, $\tau_6$ (3) \\
 & Wrist Yaw   & $q_5$, $\dot{q}_5$, $\tau_5$ (3) \\
 & Insertion   & $\tau_3$ (1) \\
 & Pitch       & $\tau_2$ (1) \\
 & Yaw         & $\tau_1$ (1) \\
\midrule
HEX10 & 6-axis F/T & $F_x$, $F_y$, $F_z$, $T_x$, $T_y$, $T_z$ (6) \\
\midrule
\multicolumn{3}{c}{\textbf{Target Data}} \\
\midrule
ATI & 3-axis Force & $F_x$, $F_y$, $F_z$ (3) \\
\bottomrule
\end{tabular}
\end{table}

Table~\ref{tbl:model_io_sources} summarizes the model inputs, which consisted of time-windowed $n$-dimensional sequences, including the \gls{dvrk} \gls{psm} joint state readouts (motor torque $\tau_i$, joint position $q_i$, and velocity $\dot{q}_i$) for the last four joints ($i=\{4,5,6,7\}$), corresponding to the wristed degrees of freedom and roll, as well as the 6-axis force $F_i$ and torque $T_i$ measurements from the HEX10. In addition, the motor torque of the \gls{psm}'s first three joints were included as inputs as these torques directly affected the forces that were applied at the end-effector. However, the velocity and position of these joints were excluded as inputs as they were directly responsible for overall positioning of the \gls{psm} shaft, but had no significant bearing on the kinematics and dynamics of the wristed degrees of freedom. This thus prevented the model from overfitting to a specific gross positioning of the end-effector in the \gls{psm} workspace.

\subsection{Baseline Models}
To benchmark our method, we implemented a three-layer \gls{fcn} network that took the same joint-state and HEX10 force/torque inputs as the transformer model. As a non-temporal baseline, the network uses only a single time step of input data to predict the three-dimensional tool tip forces in the HEX10 frame.

We also evaluated a state-of-the-art learning-based approach to estimate internal torque. Specifically, we adapted Yang et al.'s \gls{lstm}-based inverse dynamics identification network \cite{yang2025effectiveness}. Here, the external force is estimated as
\begin{equation} 
\hat{F}_{LSTM}=J^{-T}_s\left(\tau - \hat{\tau}_{LSTM}\right(\dot{q},q))\text{,}
\end{equation}
where $J_s$ denotes the \gls{psm} spatial Jacobian, $\tau$ is the measured joint torque from the \gls{dvrk}, and $\hat{\tau}_{LSTM}$ is the internal torque estimated by the model through dVRK joint velocities and positions inputs. To facilitate comparison, we transform the spatial-frame forces into the HEX10 frame as
\begin{equation}\label{eq:force_transform}
\hat{F}_{LSTM}'=\left[ \R{PSM}{Roll}(q)\ \R{Roll}{HEX10} \right]^T \hat{F}_{LSTM}\text{,}
\end{equation}
using the joint positions $q_i$ ($i \in \{1,2,3,4\}$) of the \gls{dvrk}.

In \cite{yang2025effectiveness}, the \gls{lstm} baseline was trained on free space trajectories (i.e., external forces were zero) only. To benchmark it against our method fairly, we extended training to all data, including the load space. The loss function was updated as
\begin{equation}
    \mathcal{L} = \frac{1}{n} \sum_{i=1}^{n} (F'_{ATI} - \hat{F}_{LSTM}')^2\text{,}
\end{equation}
which is the MSE loss between estimated forces and ground truth force with respect to the HEX10 frame. Its hyperparameters were adopted from the original paper. A grid search was performed for the transformer and \gls{fcn} networks, and the best hyperparameter combinations were selected based on the lowest validation loss. Specific values used for each model are listed in Table~\ref{tab:hparams_combined} of the supplementary material.

\subsection{Training and Inference Data Processing}  

We used the PyTorch Lightning framework for streamlined training and evaluation. The model was trained with the Adam optimizer and $L_2$ (MSE) loss between predicted and ground-truth forces. Regularization was applied using an $L_2$ penalty loss on all model parameters. Early stopping with a patience of 10 epochs based on validation loss was used to prevent overfitting. The learning rate decayed by 0.1 after validation loss had stopped improving for 3 epochs. After completing the last epoch, the model weights from the epoch with the lowest validation loss were chosen as the best model.

For the target data, ATI sensor forces
\begin{equation}
    F_{ATI}' = \left[ \R{ati}{PSM}\ \R{PSM}{Roll}(q)\ \R{Roll}{HEX10} \right]^T F_{ATI} \text{,}
\end{equation} 
are transformed into the HEX10 frame as in~(\ref{eq:force_transform}). To prevent bias toward a single axis, the ATI's $F_x$, $F_y$, and $F_z$ force values are normalized using the mean and standard deviation of the training set.  

During evaluation of a test set, predictions from the network are rescaled back to original ATI force range to allow for interpretable comparison against the ground-truth force. A bi-directional median filter with kernel size, $k$, of 71 data points, or 0.71 seconds was derived empirically to smooth the output.

\section{Automated Benchtop Calibration and Performance Evaluation}

\label{sec:benchtop_calib}

To calibrate and validate our combined force sensor and data-driven force estimation approach, we performed automated benchtop experiments where we programmed the \gls{dvrk} to replicate varied joint configurations, motion speeds, and tool-environment interaction scenarios to provide a diverse set of training and validation data. During these experiments, we recorded joint positions, velocities, and motor torques from the \gls{dvrk}, raw six-axis force and torque signals from the HEX10, and ground truth interaction forces from an ATI Nano17.

\begin{figure*}[h]
    \centering
    \includegraphics[width=0.9\textwidth]{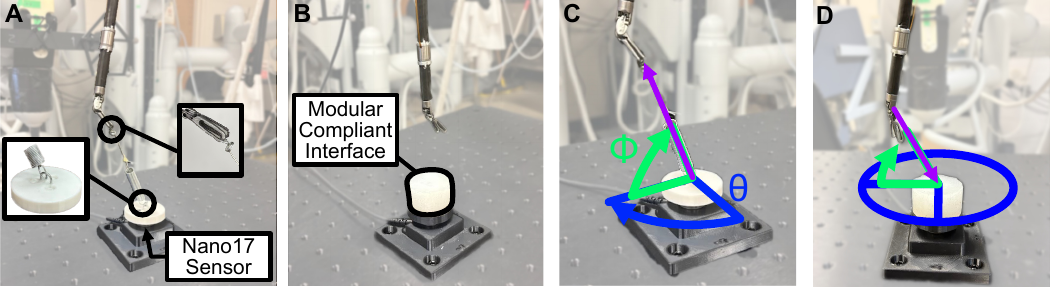}
    \caption{Automated benchtop data collection setup: 
    (A) Traction fixture. (B) Palpation fixture. (C) End-effector position defined in task space for traction fixture. (D) Task space end-effector position for palpation fixture, with approach vector defined in purple.}
    \label{fig:fixtures}
\end{figure*}

\subsection{Experimental Setup}

\subsubsection{Traction Setup}

Sampling traction data required the end-effector of the EndoWrist to be connected to the ground truth sensor such that the wristed pose of the end-effector could achieve different configurations, while under a controllable external traction load. An anchoring component was 3D printed from PLA, and attached to the Nano17, as shown in Fig.\,\ref{fig:fixtures} A. A metal ring was attached to the anchoring component to form a D-ring. This provided an attachment point for a linear spring element with a spring rate of 0.154 N/mm, which was connected to the end-effector via a nylon string with a hook on the end. With this design, the pose of the end-effector could change freely without introducing significant disturbances.

This setup allowed gross traction to be controlled by using the first three joints (outer roll, outer yaw, and insertion) of the \gls{dvrk} to alter the Cartesian position of the end-effector relative to the Nano17. The direction of the traction force could be varied by changing the orientation of a vector defined from the ground truth sensor to the wrist joint of the end-effector. The azimuth angle of this vector was defined as $\phi$ and the elevation angle, $\theta$. The magnitude of applied force could be varied by modifying the length of the vector, as this corresponded to the extension of the spring, as shown in Fig.\,\ref{fig:fixtures} A. Once the outer roll, outer yaw, and insertion values are prescribed, defining $\phi$, $\theta$, and stiffness, we iterate over a set of end-effector wrist poses, to produce internal and external force data around that local region of the input-output space.

\subsubsection{Palpation Setup}

To safely collect palpation data with different wrist poses, we introduced a compliant silicone interface between the Nano17 and the EndoWrist end-effector. Three cylindrical interfaces of different compliance were cast using a 3D-printed mold. The hardest and medium softness interfaces were cast from Zhermack Elite Double 22 Shore A and 8 Shore A silicon, respectively. The softest interface was cast from EcoFlex Shore 00-10.  As we describe below, the target palpation distance remained the same between different data collection runs, thus the forces were varied by the hardness of the interface.

The sampling process for palpation was defined similarly to traction, however, as the palpation fixture is not physically connected to the end effector as in traction, an approach and retraction motion was required. As in traction, a vector is defined from the center of the ground truth force sensor as shown in Fig.\,\ref{fig:fixtures}D. For palpation, this is defined as the approach vector. The z position of the reference point is defined 8 mm below the surface of the compliant interface. For each value of $\phi$ and $\theta$, the end effector would start at a distance of 5\,cm from the base point on the approach vector. Before each approach, a new end effector configuration is chosen. Next, the end effector moves to the base reference point along the approach vector using the \texttt{servo-cp} movement command provided by the \gls{dvrk} Python API. After completing the palpation motion, the end effector returns to its original point at the same $\phi$ and $\theta$ at the 5 cm retraction point. The configuration can then be incremented to its next state.

\subsubsection{Free Space Setup}

Before collecting the traction or palpation data, a baseline free space trajectory was recorded for each. This was performed by removing the traction or palpation interface and raising the \gls{psm} in the $z$ direction so that the end-effector could execute the prescribed motions from either trajectory without contacting any external surface. 

For validation, $\theta$ and $\phi$ were fixed at zero while the end effector executed randomized configurations over a defined time interval. This dataset served as an independent reference to evaluate model performance under simplified orientation constraints.

\begin{table}[t]
\centering
\caption{Factorial design of automated experimental data collection and number of trial repetitions per condition}
\label{tbl:design_of_experiment}
\resizebox{\linewidth}{!}{
\begin{tabular}{cccccc}
\toprule
\textbf{\makecell{Condition}} & \textbf{$\phi$} & \textbf{$\theta$} & \textbf{Force} &
\textbf{\makecell{Wrist \\ Configuration}} & 
\textbf{\makecell{Trials}} \\
\midrule
\multicolumn{6}{c}{\textbf{Training and Validation Data}} \\
\midrule
Traction                & 6 & 3 & 3 & 25 & 2 \\
Palpation            & 6 & 3 & 1 & 25 & 3 \\
Free Space (Traction)   & 6 & 3 & N/A & 25 & 1 \\
Free Space (Palpation) & 6 & 3 & N/A & 25 & 1 \\
\midrule
\multicolumn{6}{c}{\textbf{Test Data}} \\
\midrule
Traction        & 6 & 3 & 3 & 15 & 1 \\
Palpation    & 6 & 3 & 1 & 9 & 3 \\
Free Space     & 0 & 0 & N/A & \makecell{Continuous \\ random states} & 1 \\
\bottomrule
\end{tabular}}
\vspace{1em}

\caption{Automated dataset composition and split}
\label{tab:data_split}
\begin{tabular}{lcccc}
\toprule
\textbf{Dataset} &
\makecell{\textbf{Duration}\\\textbf{(min at 100\,Hz)}} &
\makecell{\textbf{Free}\\\textbf{(\%)}} &
\makecell{\textbf{Traction}\\\textbf{(\%)}} &
\makecell{\textbf{Palpation}\\\textbf{(\%)}} \\
\midrule
Train       & 131.1 & 9.5  & 42.9 & 47.6 \\
Validation  & 14.6  & 9.5  & 42.9 & 47.6 \\
Test        & 54.5  & 18.3 & 42.4 & 39.2 \\
\bottomrule
\end{tabular}
\end{table}

\subsection{Experimental Procedure}
\label{sec:procedure}

The experimental design of the data sampling procedure was configured to sample a large variation of end-effector configurations consecutively in traction and palpation across the space of possible $\phi$ and $\theta$ values. For both traction and palpation, $\phi$ was tested across values from $0^{\circ}$ to $360^{\circ}$. $\theta$ was incremented from $0^{\circ}$ to $90^{\circ}$. At a given set of $\phi$ and $\theta$, the traction force was incremented by extending the spring length by either 0, 1, or 2 cm. The palpation forces were generated by pushing against silicone bases of increasing stiffness. The data collection sequence used nested loops over $\phi$, $\theta$, force level, and wrist configuration, with a 0.4s dwell at each configuration. One full run over these combinations constituted one trial, which was repeated several times per condition. At the beginning of each trial, a reset procedure was performed to introduce noise and increase data diversity. This involved removing the instrument, re-attaching it to the \gls{psm}, and zeroing the HEX10.  The number of levels per parameter and the number of repetitions is summarized in Table~\ref{tbl:design_of_experiment}.

Additionally, the movement pattern developed to collect training, validation, and test data had to be carefully calibrated to not overload and damage the sensor. As discussed in Section\,\ref{sec:technical_challenges}, the geometry of the EndoWrist can cause forces applied at the tool tip to produce large shaft compression forces. As will be discussed in Section\,\ref{sec:overload}, this posed a limitation on the forces and interactions that could be sampled in training and test datasets, as the compression forces had to be kept below the safety limit of the HEX10 sensor.

\subsection{Training and Testing Data Split}

\begin{figure*}[t]
    \begin{minipage}{5.775in}
    \centering
    \includegraphics[width=1\linewidth]{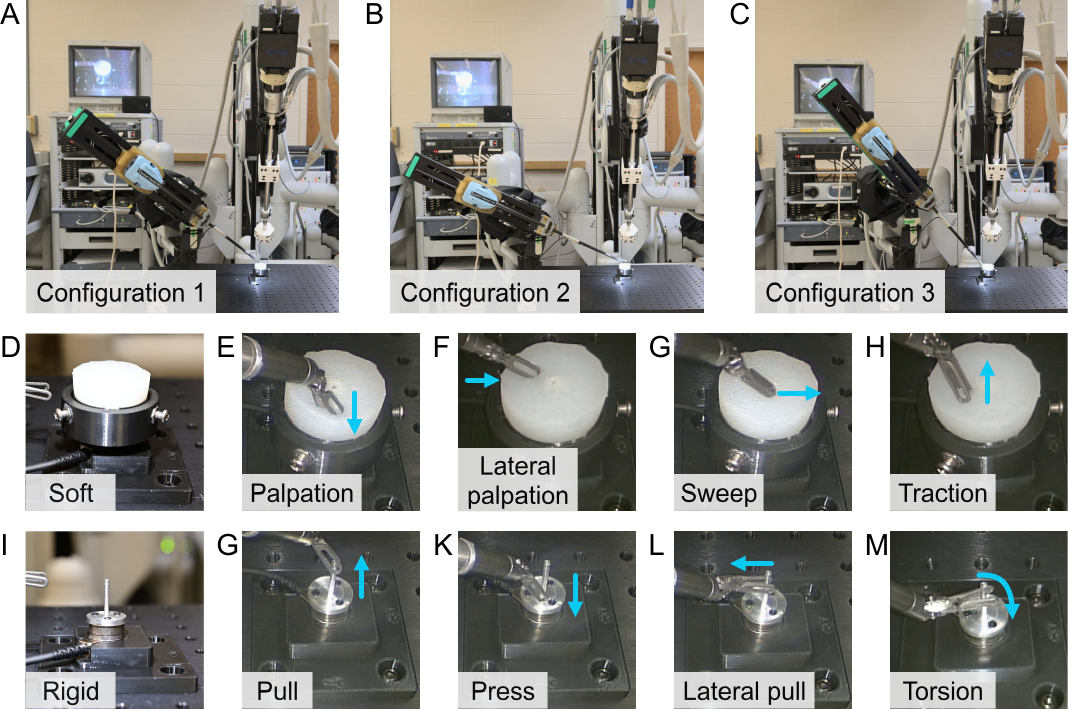}
    \end{minipage}
    \begin{minipage}{1.25in}
    \vspace{+11em}
    \caption{(A-C) PSM configurations used for evaluation under manual teleoperation. A leave-one-configuration-out cross-validation scheme was used: for each fold, demonstrations from two configurations were used for training, while the remaining configuration was held out exclusively as an unseen test set. (D-H) Movements performed in soft-contact teleoperation demonstrations. (I-M) Movements performed in rigid-contact teleoperation demonstrations.}
    \label{fig:teleop_motion_types}
    \end{minipage}
\end{figure*}

The training, validation, and test sets were generated following the design of experiment outlined in Table~\ref{tbl:design_of_experiment}. To mitigate a dataset imbalance, the free space training and validation data were randomly down-sampled to 25\% of their original size before combining with the other datasets. An overall 65.5\% / 7.3\% / 27.2\% split was achieved to form the training, validation, and test sets, respectively. The validation portion was split from the training data randomly, ensuring its distribution covers all different. Because the time required per trial differed between palpation and traction, we collected more palpation trials to balance the data distribution. Table \ref{tab:data_split} describes the final composition of each dataset.

\subsection{Evaluation Metrics}

The performance of the model was evaluated based on the \gls{rmse} and the \gls{nrmse}, 
\begin{equation}
\text{NRMSE}_{F_i} = 
\frac{\text{RMSE}_{F_i}}{\max_k F_{i,k} - \min_k F_{i,k}} \times 100 ,
\quad i \in \{x, y, z\} \text{,}
\label{eq:nrmse}
\end{equation}
which normalizes the \gls{rmse} over the measured force range of its trajectory. 
At a practical level, tracking force variations might be more informative for human and robot sensorimotor control than raw absolute predictions, hence, we also measured correlation of the predicted and ground truth forces using the coefficient of determination, $R^2$.

\section{Model Finetuning and Evaluation under Manual Teleoperation}

The automated approach described in Section\,\ref{sec:benchtop_calib} provided a large amount of systematically diversified data for calibration and validation. However, such data does not necessarily include realistic movement trajectories typically seen in teleoperation. Therefore, we collected a manual teleoperation dataset to further assess model generalization to these unseen wrist configurations and loading conditions.

We theorized that residual generalization error in teleoperation could be corrected by finetuning the pre-trained model on a limited amount of new spatial positions and teleoperated manipulations. During finetuning, all model parameters except the last decoder block were frozen. This greatly reduced the number of trainable parameters (see details in Table~\ref{tab:model_details_appendix} of the supplementary material). We adopted a leave-one-configuration-out cross-validation strategy across the three \gls{psm} configurations, resulting in three folds and thus three trained models.

While the benchtop calibration was performed with the \gls{psm} in an upright position to facilitate data collection within a larger workspace, actual teleoperation requires the \gls{psm} to be placed in configurations with larger rotations about the first two joints. 
Thus, nine demonstrations were collected at each of the three configurations shown in Fig.\,\ref{fig:teleop_motion_types}A-C. Two contact conditions, a soft (see Fig.\,\ref{fig:teleop_motion_types}D) and a rigid (see Fig.\,\ref{fig:teleop_motion_types}I) interface, were each collected as three two-minute demonstrations at each configuration. Three additional free space demonstrations were also performed via teleoperation at each configuration.

This sampling method preserves repeated observations per condition while maintaining an independent dataset in each fold. For each seen configuration, two demonstrations per condition were used for training, while one soft, one rigid, and one free space demonstration were reserved for testing on a seen configuration, and the leave-one-out configuration remained fully unseen for evaluation. For each demonstration, several specific action types were performed in random order. These are shown in Fig.\,\ref{fig:teleop_motion_types}E-H and Fig.\,\ref{fig:teleop_motion_types}G-M.

To correct offset error caused by differences in gravitational forces and small changes in internal friction between runs, a zero bias correction was applied to the model prediction. This bias was computed by averaging the first half second of predicted force for each axis for each test set, where the instrument was held stationary in free space for several seconds. The same correction was applied to the \gls{fcn}, but not to the \gls{lstm}, as it yielded no systematic performance benefit to the latter.

\section{Results and Analyses}

\subsection{Ground Truth vs. Observed Sensor Forces}\label{sec:overload}

In the test dataset, the ranges of lateral tool tip forces were approximately between $-2.9$ to $2.9$, and $-4.1$ to $3.2$ N for the x- and y-axes, respectively. Axial forces in the z-axis were higher in traction, and had a range of $-4.1$ to 4.6\,N.

\begin{figure}[t]
    \centering
    \includegraphics[width=\linewidth]{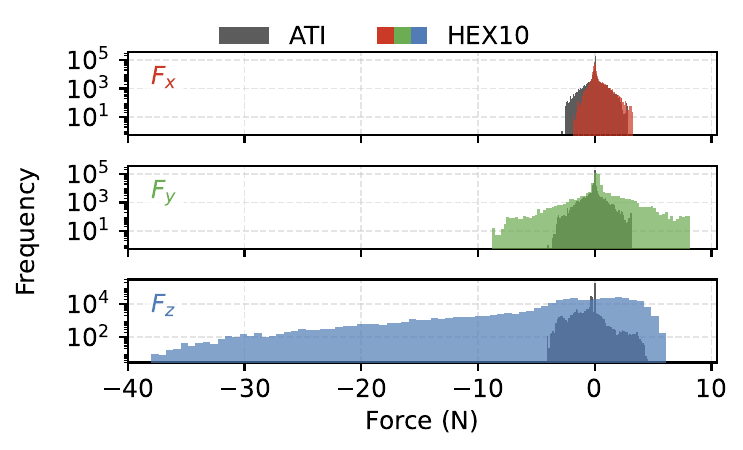}
    \vspace{-2em}
    \caption{Histograms comparing ground truth vs. HEX10 forces for test dataset.}
    \label{fig:hex10_vs_ati_hist}
\end{figure}

The range of forces observed by the distal HEX10 sensor in the test dataset had significantly higher magnitude than the forces applied at the tool tip, as illustrated in Fig.\,\ref{fig:hex10_vs_ati_hist} histograms. The high internal forces reduced both sensor accuracy and its effective sensing range. These can be described in relation to the sensor's nominal range and overload limit of the HEX10 sensor, which are $\pm$25\,N, and $\pm$100\,N, respectively. 

The observed pretension force on the HEX10 sensor during cable calibration was 6 N. Thus, with this initial compression, the sensing range was reduced to $-19$ to $+$25\,N, while the overload range was reduced to $-94$ to $+$100\,N. As shown in Fig.\,\ref{fig:hex10_vs_ati_hist}, the compression force observed in the test data came within\,15 N of the overload limit, and was beyond the nominal range. This likely influenced the force estimation accuracy in configurations with high internal forces. 

\begin{figure*}[t]
    \centering
    \includegraphics[width=1\linewidth]{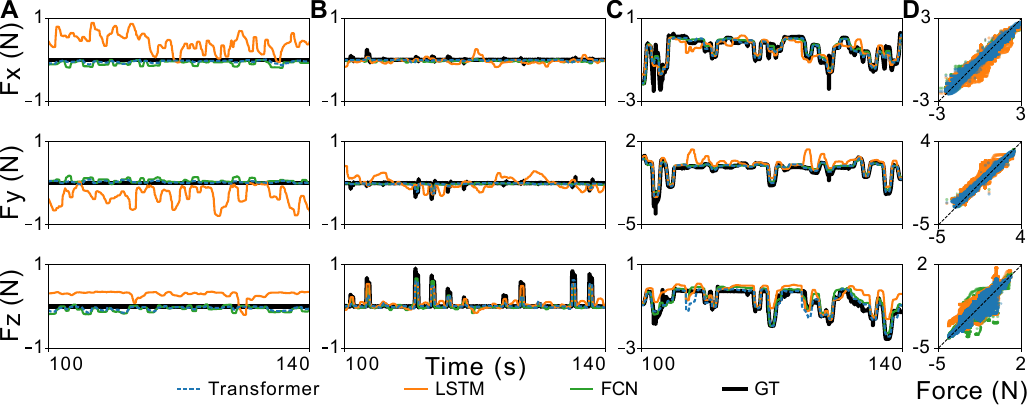}
    \vspace{-2em}
    \caption{Model predicted vs ground truth force for representative sections of (A) free, (B) palpation, and (C) traction benchtop performance evaluation runs. (D) Predicted force vs ground truth force per axis for the overall benchtop evaluation dataset.}
    \label{fig:plotted_test_data}
\end{figure*}

\begin{table*}[t]
\centering
\caption{System performance comparison on the Automated test set (best NRMSE per axis in bold)}
\label{tab:automated_results}
\begin{tabular}{c c ccc ccc ccc}
\toprule
\multirow{2}{*}{Axis} & \multirow{2}{*}{Range (\si{\newton})}
& \multicolumn{3}{c}{Transformer (Ours)}
& \multicolumn{3}{c}{LSTM~\cite{yang2025effectiveness}}
& \multicolumn{3}{c}{FCN} \\
\cmidrule(lr){3-5}\cmidrule(lr){6-8}\cmidrule(lr){9-11}
& & RMSE (\si{\newton}) & NRMSE (\%) & $\mathbf{R^2} (\%)$ & RMSE (\si{\newton}) & NRMSE (\%) & $\mathbf{R^2}$ (\%) & RMSE (\si{\newton}) & NRMSE (\%) & $\mathbf{R^2}$ (\%) \\
\midrule
$F_x$ & $5.73$ & $0.08$ & $\mathbf{1.45}$ & $96.25$ & $0.23$ & $4.04$ & $70.81$ & $0.11$ & $1.93$ & $93.36$ \\
$F_y$ & $7.22$ & $0.09$ & $\mathbf{1.31}$ & $96.08$ & $0.32$ & $4.39$ & $56.23$ & $0.11$ & $1.59$ & $94.28$ \\
$F_z$ & $8.69$ & $0.20$ & $\mathbf{2.26}$ & $94.03$ & $0.35$ & $4.05$ & $80.89$ & $0.26$ & $2.99$ & $89.60$ \\
\bottomrule
\end{tabular}
\end{table*}

\subsection{Benchtop Performance Evaluation}\label{sec:benchtop_performance}

The \gls{rmse} and \gls{nrmse} for each axis on the test dataset are presented in Table~\ref{tab:automated_results}. Our proposed force estimation approach achieved consistently high predictive accuracy across all axes in the automated test dataset compared with the other two methods. In the lateral x- and y-axes, it achieved an \gls{rmse} under 0.1\,N, with \gls{nrmse} below $2\%$. Axial z-axis force estimation achieved slightly lower accuracy, with \gls{rmse} of 0.2\,N and \gls{nrmse} under $3\%$. The $R^2$ values for each axis are also listed in Table~\ref{tab:automated_results}, and exceeded 0.9 for all axes, indicating that the system closely tracked force variations.

The \gls{lstm}-based free space torque estimator adapted from \,\cite{yang2025effectiveness} predicted forces with \gls{rmse} under 0.4\,N on all axes, which is below the values reported for the dVRK Classic in the original paper. Due to the lower force range of our dataset, the \gls{nrmse} of the LSTM was higher than their prior results. The \gls{fcn} showed lower accuracy than our transformer model on all three axes.

The above results are qualitatively illustrated in Fig.\,\ref{fig:plotted_test_data}\,A-C, which depict free space, palpation, and traction force trajectories from the test dataset, respectively. In free space, the system was able to estimate zero tool tip forces within a $\pm0.057$\,N \gls{rmse} noise envelope in the x- and y-axes, and $\pm0.072$\,N in the z-axis. In palpation, it lost accuracy at high force levels, which was consistent with our observations of sensor overload as discussed in Section\,\ref{sec:overload}. The most degraded performance was seen in traction. 
The x- and y-axes show mismatches when the lateral force changes fast, and hard to fit workload peaks. The z-axis estimates were more susceptible to error at higher force levels. This performance reduction was expected under the aforementioned overload assumption, as traction forces applied at the tool tip would require much higher internal cable forces to hold a particular wrist pose. This performance is qualitatively illustrated in Fig.\,\ref{fig:plotted_test_data}D, which shows how the force estimates linearly tracked lateral force variations closely, while estimates in the z-direction were more variable, and sometimes displayed a saturation behavior.

\subsection{Model Input Ablations}\label{sec:model_input_ablations}

\begin{figure*}[t]
    \centering
    \includegraphics[width=1\linewidth]{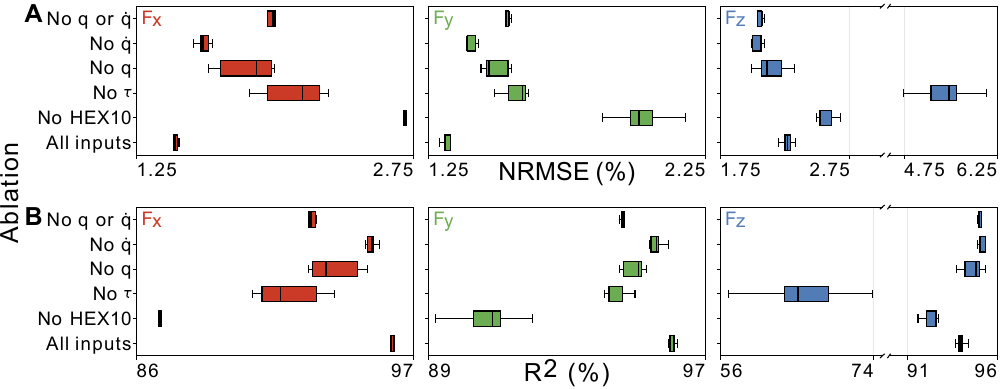}
    \vspace{-2em}
    \caption{Ablation comparison across force components on the automated benchtop dataset. For each ablation group (y-axis), horizontal box plots summarize the distributional statistics across all runs, for (A) NRMSE, and (B) $R^2$.}
    \label{fig:ablation_results}
\end{figure*}

Input ablations were performed to evaluate the contribution of each sensor modality, i.e. no HEX10 force/torque $F_i$ and $T_i$, no motor torques $\tau_i$, no joint positions $q_i$, no joint velocities $\dot{q}_i$, or no joint positions and velocities, to overall performance. This was done by removing the corresponding input, and refitting a force estimation model from scratch.  Our results indicate that the force estimation model relied most heavily on the distal sensor measurements to estimate lateral force components, while axial forces were more closely correlated with \gls{psm} actuator torques, particularly those of the insertion axis $\tau_3$. As shown in Fig.\,\ref{fig:ablation_results}, when the force and torque data from the distal HEX10 sensor were removed, the $R^2$ performance in the x- and y-axes decreased, while the \gls{nrmse} increased. However, the $R^2$ and \gls{nrmse} in the z-axis were not significantly affected. The opposite was true when the motor torques of the \gls{psm} were ablated. Finally, the removal of \gls{psm} joint position and velocity data from the input minimally impacted x- and y-axes accuracy.

\subsection{Teleoperated Performance Evaluation}\label{sec:teleoperated_performance}
\begin{figure*}[t]
    \centering
    \includegraphics[width=1\linewidth]{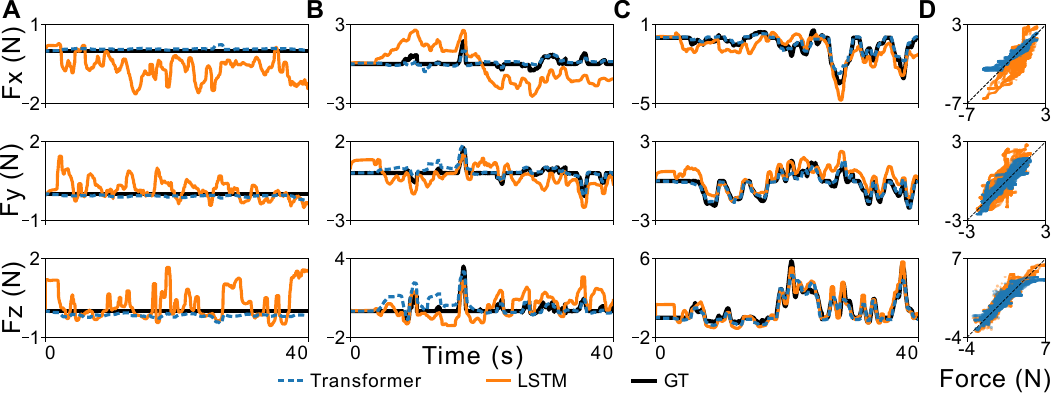}
    \vspace{-2em}
    \caption{Model predicted vs ground truth force for representative sections of (A) free, (B) rigid, and (C) soft contact conditions teleoperation performance evaluation runs. (D) Predicted force vs ground truth force per axis on the test sets shown in (B) and (C).}
    \label{fig:teleop_time_plot}
\end{figure*}

The metrics of our cross-validation were averaged over all trained models for both transformer and \gls{lstm} methods and are summarized in Table~\ref{tab:transfer_learning_results}. Model predictions from an illustrative unseen configuration on each contact condition are shown in Fig.\,\ref{fig:teleop_time_plot}A-C.

After finetuning, our method achieved a \gls{rmse} below 0.2\,N in the x- and y-axes, and 0.4 N in the z-axis across all conditions tests. Linear tracking performance was poorer than on the data-rich automated test set, with $R^2$ = 0.8. The soft contact condition showed a slight decrease in \gls{nrmse} compared to the rigid contact, with an error of approximately 5\%. The model generalized well under unseen configurations across all conditions, with \gls{nrmse} below 6\% in all cases.

The \gls{lstm} was able to achieve \gls{rmse} accuracy of below 0.85\,N which is close to what was reported in \cite{yang2025effectiveness}. However, it exhibited poorer accuracy on unseen configurations with its $R^2$ in the x direction noticeably reduced due to a prediction offset as illustrated for the soft contact scenario in Fig.\,\ref{fig:teleop_time_plot}C. Unlike with the transformer and the \gls{fcn}, the offset was not systematically correctable by applying an offset bias.

\begin{table*}[t]
\centering
\caption{Finetune performance comparison across teleoperation dataset \\ (mean $\pm$ std all cross-validation models; min-max normalized color intensity: green=better, red=worse)}
\label{tab:transfer_learning_results}
\begin{tabular}{c c c c ccc ccc}
\toprule
\multirow{2}{*}{Condition} & \multirow{2}{*}{Config} & \multirow{2}{*}{Axis} & \multirow{2}{*}{Range (\si{\newton})}
& \multicolumn{3}{c}{Transformer (Ours)}
& \multicolumn{3}{c}{LSTM~\cite{yang2025effectiveness}} \\
\cmidrule(lr){5-7}\cmidrule(lr){8-10}
& & & & RMSE (\si{\newton}) & NRMSE (\%) & $\mathbf{R^2}$ (\%) & RMSE (\si{\newton}) & NRMSE (\%) & $\mathbf{R^2}$ (\%) \\
\midrule
\multirow{6}{*}{Rigid} & \multirow{3}{*}{Seen} & $F_x$ & $3.63 \pm 0.35$ & \teleopcolorcell{67BF7B}{0.13}{0.02} & \teleopcolorcell{63BE7B}{3.60}{0.31} & \teleopcolorcell{64BE7B}{90.28}{4.17} & \teleopcolorcell{AED37F}{0.29}{0.02} & \teleopcolorcell{AED37F}{8.47}{1.14} & \teleopcolorcell{7BC57C}{47.09}{23.58} \\
 &  & $F_y$ & $2.78 \pm 0.15$ & \teleopcolorcell{63BE7B}{0.11}{0.02} & \teleopcolorcell{6CC07B}{4.25}{0.49} & \teleopcolorcell{64BE7B}{88.09}{3.34} & \teleopcolorcell{9BCE7E}{0.25}{0.01} & \teleopcolorcell{B7D67F}{9.01}{0.53} & \teleopcolorcell{7BC57C}{47.10}{1.89} \\
 &  & $F_z$ & $6.70 \pm 1.11$ & \teleopcolorcell{AED37F}{0.29}{0.04} & \teleopcolorcell{70C17B}{4.47}{0.96} & \teleopcolorcell{69BF7B}{80.71}{3.00} & \teleopcolorcell{E4E382}{0.42}{0.04} & \teleopcolorcell{8FCA7D}{6.46}{1.68} & \teleopcolorcell{75C37C}{58.64}{12.30} \\
\cmidrule(lr){2-10}
 & \multirow{3}{*}{Unseen} & $F_x$ & $3.52 \pm 0.69$ & \teleopcolorcell{6EC17B}{0.14}{0.04} & \teleopcolorcell{6AC07B}{4.09}{0.54} & \teleopcolorcell{66BF7B}{85.71}{7.05} & \teleopcolorcell{F8696B}{0.84}{0.60} & \teleopcolorcell{F8696B}{23.44}{12.14} & \teleopcolorcell{F8696B}{-475.12}{640.21} \\
 &  & $F_y$ & $2.73 \pm 0.55$ & \teleopcolorcell{63BE7B}{0.11}{0.03} & \teleopcolorcell{6BC07B}{4.19}{0.28} & \teleopcolorcell{64BE7B}{88.30}{4.35} & \teleopcolorcell{E2E282}{0.41}{0.14} & \teleopcolorcell{FDD57F}{15.19}{4.16} & \teleopcolorcell{BED880}{-74.78}{108.26} \\
 &  & $F_z$ & $6.67 \pm 0.34$ & \teleopcolorcell{B1D47F}{0.30}{0.02} & \teleopcolorcell{71C27B}{4.54}{0.48} & \teleopcolorcell{69BF7B}{80.95}{2.41} & \teleopcolorcell{FA9673}{0.71}{0.14} & \teleopcolorcell{D6DF81}{10.89}{1.72} & \teleopcolorcell{9DCE7E}{-14.62}{20.64} \\
\midrule
\multirow{6}{*}{Soft} & \multirow{3}{*}{Seen} & $F_x$ & $4.45 \pm 0.70$ & \teleopcolorcell{7DC57C}{0.18}{0.04} & \teleopcolorcell{69BF7B}{4.05}{0.26} & \teleopcolorcell{63BE7B}{92.50}{1.75} & \teleopcolorcell{A9D27F}{0.28}{0.04} & \teleopcolorcell{90CB7D}{6.48}{0.19} & \teleopcolorcell{69BF7B}{80.83}{2.83} \\
 &  & $F_y$ & $3.94 \pm 0.36$ & \teleopcolorcell{7AC47C}{0.17}{0.04} & \teleopcolorcell{6EC17B}{4.35}{0.57} & \teleopcolorcell{63BE7B}{91.90}{2.61} & \teleopcolorcell{A0CF7E}{0.26}{0.02} & \teleopcolorcell{93CC7D}{6.75}{0.76} & \teleopcolorcell{69BF7B}{81.02}{3.77} \\
 &  & $F_z$ & $6.80 \pm 0.85$ & \teleopcolorcell{C3D980}{0.34}{0.04} & \teleopcolorcell{79C47C}{5.01}{0.15} & \teleopcolorcell{65BE7B}{86.51}{2.20} & \teleopcolorcell{B1D47F}{0.30}{0.01} & \teleopcolorcell{70C17B}{4.47}{0.62} & \teleopcolorcell{64BE7B}{88.75}{3.45} \\
\cmidrule(lr){2-10}
 & \multirow{3}{*}{Unseen} & $F_x$ & $4.51 \pm 0.37$ & \teleopcolorcell{84C77C}{0.19}{0.07} & \teleopcolorcell{6CC07B}{4.24}{1.19} & \teleopcolorcell{64BE7B}{89.93}{4.49} & \teleopcolorcell{FBA676}{0.67}{0.28} & \teleopcolorcell{FDD880}{14.97}{5.66} & \teleopcolorcell{B4D57F}{-56.72}{98.90} \\
 &  & $F_y$ & $3.89 \pm 0.79$ & \teleopcolorcell{7CC57C}{0.18}{0.04} & \teleopcolorcell{71C27B}{4.59}{0.76} & \teleopcolorcell{63BE7B}{91.01}{1.72} & \teleopcolorcell{FEE182}{0.50}{0.15} & \teleopcolorcell{FEE482}{14.04}{6.67} & \teleopcolorcell{8CCA7D}{16.62}{48.57} \\
 &  & $F_z$ & $6.16 \pm 0.90$ & \teleopcolorcell{BFD880}{0.33}{0.04} & \teleopcolorcell{80C67C}{5.50}{0.70} & \teleopcolorcell{66BF7B}{84.88}{6.80} & \teleopcolorcell{FEEA83}{0.47}{0.10} & \teleopcolorcell{A5D17E}{7.84}{0.68} & \teleopcolorcell{6FC17B}{69.53}{7.77} \\
\bottomrule
\end{tabular}
\end{table*}

\subsection{Inference Speed}\label{sec:inference_speed}

As shown in Table~\ref{tab:inference_perf} the transformer model averages 499\,Hz inference speed. Although below the gold-standard 1\,kHz recommended for bilateral teleoperation, such speed is sufficient for compliant interactions with deformable tissues. Although the \gls{fcn} is faster than 1\,kHz, its lower estimation accuracy (Table~\ref{tab:automated_results}) weakens its practical utility.

\section{Discussion}

In the above experiments, we compared our data-driven temporal force estimation approach integrating distal force measurements, to a non-temporal variant, as well as a data-driven temporal model for force estimation that used only proximal joint torque estimates. Our method exhibited comparatively higher accuracy and linear tracking performance against these baselines. These demonstrate the usefulness of both the inline force sensor and time series data-driven modeling of the internal dynamics in improving force estimation.

Considering its applicability as a source for haptic feedback, our transformer approach has an NRMSE below the average Weber fraction of 12.5\%, in all force directions. This suggests that the estimation errors are likely below the human perceptual threshold and therefore provide sufficient fidelity for haptic feedback applications. However, one drawback of our current system is its limited usable force range of $\leq$6\,N. This makes it mainly suitable for estimating forces in palpation (axial compression $<$2\,N \cite{applied_surgury_forces}) and dissection (peak forces $<$5\,N \cite{wagner2002bluntdissection}).Cutting and suturing tasks require higher usable force ranges as axial tension forces are $>$10\,N \cite{applied_surgury_forces}. Here, the \gls{lstm} baseline's higher usable force range of $>\pm$10\,N makes it more practical, despite its lower RMSE accuracy, because NRMSEs in relation to the Weber Fraction become acceptable.

The deliberate omission of the positions of the first three \gls{psm} joints from our model was intended to prevent it from overfitting to a workspace configuration. However, for surgical tasks involving a larger workspace, the absence of these joint inputs may limit the model's ability to compensate for gravity across various poses. Moreover, the gravity compensation provided by the HEX10's ``tare" function and the bias correction are effective only within a limited workspace.

Relatedly, the overall influence of the above unmodeled dynamics is still limited, given the low weight of the added HEX10 sensor and interface (approximately 3.05\,g). This appears small compared with the combined weight of 36.67\,g for the original shaft and EndoWrist. This additional distal mass may nonetheless negatively affect the accuracy of the \gls{dvrk} position controllers and warrants further investigation.

In developing the data-driven force estimation model, we collected a custom dataset using a systematic automated sweep of the end-effector configurations in free space and under load. This was then augmented with manual teleoperation data. Given the promising results in manual teleoperation after finetuning, it is possible that a larger and more diverse dataset, either containing automated or human-demonstrated sampling of the input-output space, would produce more accurate force estimates. 

As discussed in Section\,\ref{sec:technical_challenges}, forces applied perpendicular to the tip of the EndoWrist jaw result in large axial compression forces due to the ratio of the jaw length to the radius of the distal control pulleys. Because of this, an instrument with a shorter jaw length would likely experience smaller resultant compression forces. For the sensor integration method developed in this work, this could result in improved signal to noise ratio. Therefore, we expect that utilizing our modeling approach with the HEX10 sensor integrated into a needle driver, with a shorter jaw length than the Cadiere Forceps, would yield a higher usable force range and better accuracy.

Based on the input ablation results in Fig.\,\ref{fig:ablation_results}, the HEX10's sensing is utilized heavily for lateral force estimation.Given the sensor's distal location, our approach will be more robust to noise from friction induced by the laparoscopic port compared to the baseline methods which utilize only proximal torque measurements. However, our approach does rely heavily on insertion-axis torque to estimate z-axis forces. This can be mitigated using a trocar force-sensing approach~\cite{fontanelli2017}, incorporating a data-driven disturbance estimator~\cite{yang2024hybrid}, or introducing a proximal sensor to cancel internal disturbances~\cite{resense_differential_force_compensation}.

The reproducibility of the developed model is impacted by the initial data-intensive calibration stage. Fitting our models under load required an external benchtop setup with a ground-truth force sensor, utilizing more than 3 hours of data collected at 100\,Hz. In contrast, the \gls{lstm} approach was developed and validated using free-space trajectories \cite{wu2021robot, yang2025effectiveness}. Such data are easier to collect, but come at the expense of lower model prediction accuracy compared to those fitted with load data. Future development will investigate how including an in-line sensor improves such free space calibration approaches.

In this work, model performance was evaluated using engineering metrics ($R^2$, \gls{nrmse}, and \gls{rmse}) on both automated and teleoperated datasets. Future work will define task-relevant metrics that better quantify how estimated force supports downstream autonomy objectives, such as stiffness discrimination and event detection in force-informed manipulation.

\begin{table}[t]
\centering
\caption{Inference performance across models and devices}
\label{tab:inference_perf}
\begin{tabular}{lccc}
\toprule
\textbf{Model} & \textbf{Device} & \textbf{End-to-End (ms)} & \textbf{End-to-End (Hz)} \\
\midrule
Transformer & GPU & 2.12 $\pm$ 0.55 & 499.66 $\pm$ 110.98 \\
LSTM~\cite{yang2025effectiveness}& GPU & 2.33 $\pm$ 0.41 & 440.54 $\pm$ 66.79 \\
FCN         & CPU  & 0.12 $\pm$ 0.04 & 8811.45 $\pm$ 1430.48 \\
\bottomrule
\end{tabular}%
\end{table}

Finally, our sensor meets our accessibility criteria set out in Section\,\ref{sec:design_obj}. Each machined shaft interface costs \$6.48 to fabricate via third-party outsourcing, while modification jigs, and calibration equipment were 3D-printed for $<$\$100 total, therefore satisfying our key cost criteria. Subsequent modification steps were performed with hand tools. The highest cost was manufacturing time. An experienced research team member fabricated the tools in less than 4 hours, which provides a reproducible lower bound for manufacturing time. Given that the manufacturing process introduces variability, moving forward, we will aim to rigorously quantify the robustness of our approach's performance over multiple tool modifications.

\section{Conclusions}

In this paper, we presented a method to integrate a tubular 6-axis force sensor into the distal end of a laparoscopic telesurgical instrument. In its current form, the proposed sensor integration method is reproducible and accessible, facilitating its use with \gls{ramis} tools that are compatible with research platforms such as the \gls{dvrk}~\cite{dvrk_main} or Raven~\cite{raven}. By integrating a data-driven dynamics estimator to compensate for internal forces in the shaft, the six-axis force sensor measurement enables accurate estimation of end-effector forces. 
A key limitation of the current system is the restricted usable force range due to amplified internal cable forces, which may limit performance in higher-force surgical tasks such as suturing.
Nonetheless, our approach facilitates the study of increasingly pertinent topics in \gls{ramis}, including haptic feedback~\cite{vuong2025effects}, automated skill assessment~\cite{hung_development_2018}, and autonomous or semi-autonomous tasks execution~\cite{haptic_palpation}, thus providing a versatile platform for advancing research in surgical tool performance and control.

\section*{Acknowledgment}
The authors acknowledge Connor Kiernan for his contribution to initial testing of interface prototypes; Kaiwen Zuo for his assistance with the integration of the sensor streams into the ROS 2 architecture. The authors acknowledge Resense GmbH's loan HEX10 sensor for the research. Compute was provided by the High Performance Computing Resource in the Core Facility for Advanced Research Computing at Case Western Reserve University. 

\bibliographystyle{IEEEtran}
\bibliography{references}

\clearpage
\onecolumn
\renewcommand{\thetable}{S\arabic{table}}

\section*{Supplementary Material}

\subsection*{Hyperparameters}\label{sec:hyperparameter}

The training batch size for all models is set to 256.

\begin{table}[h]
\centering
\caption{Hyperparameters for each model.}
\label{tab:hparams_combined}
\begin{tabular}{lccccc}
\toprule
\textbf{Model} & \textbf{LR} & \makecell{\textbf{Weight}\\\textbf{Decay}} & \makecell{\textbf{Seq}\\\textbf{Length}} & \textbf{Stride} & \makecell{\textbf{Max}\\\textbf{Epochs}} \\
\midrule
\multicolumn{6}{c}{\textbf{Automated}} \\
\midrule
Transformer & 1e-3 & 1e-4 & 100  & 5 & 100 \\
LSTM        & 5e-3 & 1e-6 & 1000 & 5 & 200 \\
FCN         & 1e-3 & 1e-4 & 1    & 5 & 100 \\
\midrule
\multicolumn{6}{c}{\textbf{Teleoperation}} \\
\midrule
Transformer & 1e-3 & 1e-4 & 100  & 5 & 100 \\
LSTM        & 1e-3 & 1e-4 & 1000 & 5 & 200 \\
FCN         & 1e-3 & 1e-4 & 1    & 5 & 100 \\
\bottomrule
\end{tabular}
\end{table}

\subsection*{Model Details}\label{sec:model_arch}

\begin{table}[h]
\centering
\caption{Model architecture details and trainable parameter counts.}
\label{tab:model_details_appendix}
\begin{tabular}{lcccc}
\toprule
\textbf{Model} & \makecell{\textbf{Hidden}\\\textbf{Units}} & \textbf{Layers} & \makecell{\textbf{Full}\\\textbf{Params}} & \makecell{\textbf{Finetune}\\\textbf{Params}} \\
\midrule
Transformer & 64  & 3 & 155K & 4.4K \\
LSTM        & 128 & 1 & 486K & 49.9K \\
FCN         & 64  & 3 & 5.8K & 0.2K \\
\bottomrule
\end{tabular}
\end{table}

\newpage
\subsection*{Experiment Summary}

\begin{table*}[h]
\centering
\caption{Summary of experiments, evaluation metrics, and corresponding objectives.}
\label{tab:experiment_summary}
\resizebox{\textwidth}{!}{
\begin{tabular}{>{\raggedright\arraybackslash}p{2.7cm} >{\raggedright\arraybackslash}p{2.7cm} >{\raggedright\arraybackslash}p{4.0cm} >{\raggedright\arraybackslash}p{5.0cm}}
\toprule
\textbf{Experiment} & \textbf{Dataset} & \textbf{Evaluation Metrics} & \textbf{Purpose} \\
\midrule
Automated benchtop performance evaluation 
& Automated benchtop dataset (traction, palpation, free space) 
& RMSE, NRMSE, $R^2$ 
& Validate force estimation under controlled traction, palpation, and free-space conditions, and compare against baseline models. \\
\addlinespace[2pt]

Model input ablations 
& Automated benchtop dataset (ablation variants) 
& RMSE, NRMSE, $R^2$ 
& Quantify the contribution of HEX10 measurements, motor torques, joint positions, and joint velocities to prediction accuracy. \\
\addlinespace[2pt]

Teleoperated performance evaluation 
& Teleoperation dataset (seen/unseen; rigid, soft, free space) 
& RMSE, NRMSE, $R^2$  
& Evaluate generalization to manual teleoperation and assess the impact of finetuning on realistic trajectories. \\
\addlinespace[4pt]

Inference speed evaluation 
& N/A 
& End-to-end latency 
& Verify real-time feasibility of the proposed model. \\
\bottomrule
\end{tabular}
}
\end{table*}


\end{document}